\def\year{2021}\relax
\documentclass[letterpaper]{article} 
\usepackage{aaai21}  
\usepackage{times}  
\usepackage{helvet} 
\usepackage{courier}  
\usepackage[hyphens]{url}  
\usepackage{graphicx} 
\usepackage{natbib}  
\usepackage{caption} 
\usepackage[switch]{lineno}  %
\newcommand*\patchAmsMathEnvironmentForLineno[1]{
  \expandafter\let\csname old#1\expandafter\endcsname\csname #1\endcsname
  \expandafter\let\csname oldend#1\expandafter\endcsname\csname end#1\endcsname
  \renewenvironment{#1}
     {\linenomath\csname old#1\endcsname}
     {\csname oldend#1\endcsname\endlinenomath}}
\newcommand*\patchBothAmsMathEnvironmentsForLineno[1]{
  \patchAmsMathEnvironmentForLineno{#1}
  \patchAmsMathEnvironmentForLineno{#1*}}
\AtBeginDocument{
\patchBothAmsMathEnvironmentsForLineno{equation}
\patchBothAmsMathEnvironmentsForLineno{align}
\patchBothAmsMathEnvironmentsForLineno{flalign}
\patchBothAmsMathEnvironmentsForLineno{alignat}
\patchBothAmsMathEnvironmentsForLineno{gather}
\patchBothAmsMathEnvironmentsForLineno{multline}
}

\usepackage[T1]{fontenc}
\usepackage{algorithm}
\usepackage{algorithmic}
\usepackage{subfigmat}
\usepackage{amssymb,amsmath,amsthm,amsfonts}
\usepackage{svg}
\usepackage{here}

\newcommand{\Tbref}[1]{Table~\ref{#1}}

\newcommand{\Figref}[1]{Figure~\ref{#1}}

\newcommand{\Algref}[1]{Algorithm~\ref{#1}}

\newcommand{\E}{\mathbb{E}}

\newcommand{\memo}[1]{\textcolor{red}{#1}}

\newcommand{\comm}[1]{}

\graphicspath{
{./figure/}
}

\title{Off-Policy Meta-Reinforcement Learning Based on Feature Embedding Spaces}

\author{

        Takahisa Imagawa,\textsuperscript{\rm 1}
        Takuya Hiraoka, \textsuperscript{\rm 123}
        Yoshimasa Tsuruoka \textsuperscript{\rm 14} \\
}
\affiliations{
    \textsuperscript{\rm 1} National Institute of Advanced Industrial Science and Technology \\
    \textsuperscript{\rm 2} NEC Central Research Laboratories \\
    \textsuperscript{\rm 3} RIKEN Center for Advanced Intelligence Project  \\
    \textsuperscript{\rm 4} The University of Tokyo, Tokyo  \\
    imagawa.t@aist.go.jp, takuya-h1@nec.com, tsuruoka@logos.t.u-tokyo.ac.jp
}

\comm{
  \begin{icmlauthorlist}
\icmlauthor{Takahisa Imagawa}{aist}
\icmlauthor{Takuya Hiraoka}{aist,nec,aip}
\icmlauthor{Yoshimasa Tsuruoka}{aist,ut}
\end{icmlauthorlist}

\icmlaffiliation{aist}{National Institute of Advanced Industrial Science and Technology, Tokyo, Japan}
\icmlaffiliation{nec}{NEC Central Research Laboratories, Kanagawa, Japan}
\icmlaffiliation{aip}{RIKEN Center for Advanced Intelligence Project, Tokyo, Japan}
\icmlaffiliation{ut}{The University of Tokyo, Tokyo, Japan}

\icmlcorrespondingauthor{Takahisa Imagawa}{imagawa.t@aist.go.jp}
\icmlkeywords{Reinforcement Learning, Meta-Learning, Transfer, Amortized Inference}
・Takahisa Imagawa ( National Institute of Advanced Industrial Science and Technology) <imagawa.t@aist.go.jp>
・Takuya Hiraoka ( NEC Corporation) <takuya-h1@nec.com>
・Yoshimasa Tsuruoka ( The University of Tokyo) <tsuruoka@logos.t.u-tokyo.ac.jp>
}

\begin{document}
\maketitle

\begin{abstract}

  \comm{  
    Meta-reinforcement learning (RL) addresses the problem of sample inefficiency in deep RL by using experience obtained in past tasks for a new task to be solved.
    However, most meta-RL methods require partially or fully on-policy data, i.e., they cannot reuse the data collected by past policies, which hinders the improvement of sample efficiency.  
    To alleviate this problem, we propose a novel off-policy meta-RL method, \textit{embedding learning and evaluation of uncertainty} (ELUE). 
    An ELUE agent is characterized by the learning of a shared feature embedding space among tasks.
    It learns a belief model over the embedding space and a belief conditional policy and Q-function. 
    \memo{Then, for a new task, it collects data by the pretrained policy, and updates belief based on the belief model. The performace can be improved with small amount of data by the belief update.}
    \memo{In addition, it updates parameters of the neural networks for modifying the pretrained relationships if it earns enough data.}
    We show that our proposed method outperforms existing methods through experiments with a meta-RL benchmark.
    }

    Meta-reinforcement learning (RL) addresses the problem of sample inefficiency in deep RL by using experience obtained in past tasks for a new task to be solved.
    However, most meta-RL methods require partially or fully on-policy data, i.e., they cannot reuse the data collected by past policies, which hinders the improvement of sample efficiency. 
    To alleviate this problem, we propose a novel off-policy meta-RL method, \textit{embedding learning and evaluation of uncertainty} (ELUE). 
    An ELUE agent is characterized by the learning of a feature embedding space shared among tasks.
    It learns a belief model over the embedding space and a belief-conditional policy and Q-function. 
    Then, for a new task, it collects data by the pretrained policy, and updates its belief based on the belief model.
    Thanks to the belief update, the performance can be improved with a small amount of data. 
    In addition, it updates the parameters of the neural networks to adjust the pretrained relationships when there are enough data.
    We demonstrate that ELUE outperforms state-of-the-art meta RL methods through experiments on meta-RL benchmarks.
\end{abstract}

\section{Introduction}
Deep reinforcement learning (DRL) has shown superhuman performance in several domains, such as computer games and board games~\cite{silver2017mastering, berner2019dota}.
However, conventional DRL considers only learning for a single task and does not reuse experience from past tasks.
This is a major reason for the sample inefficiency in conventional DRL.

To overcome this problem, the concept of meta-learning has been proposed~\cite{schmidhuber1996simple}.
Meta-learning is a class of methods for learning how to efficiently learn with a small amount of data on a new task by utilizing previous experience.
An instance of meta-learning consists of two phases: meta-training and meta-testing.
In meta-training, the agent prepares itself for learning in meta-testing by using some training tasks.
In meta-testing, the agent is evaluated on the basis of its performance on the task to be solved. 
Although meta-learning aims to improve sample efficiency in meta-testing, the sample efficiency in meta-training is also important from the perspective of computational cost~\cite{mendonca2019guided, rakelly2019efficient}.  

One type of meta-learning methods is called {\it gradient-based meta-learning}, and they include MAML~\cite{finn2017model} and Reptile~\cite{nichol2018first}. 
These methods learn to reduce the loss of the model whose parameters (e.g., weights of neural networks) are updated over several steps.
\citet{finn2017model} have shown that the sample efficiency of MAML in meta-testing is improved compared with that of naive pretraining methods.
However, most reinforcement learning (RL) applications of these methods work with on-policy methods~\cite{mendonca2019guided}, which are less sample efficient than off-policy methods because on-policy methods cannot reuse the data collected by old policies.
In addition, the performance of the learned initial parameters in meta-testing can be very poor in some cases until the parameters are updated.  
For example, there are tasks where an agent aims to reach a goal as fast as possible, and the tasks differ in terms of goal positions.
Let us assume that there are two tasks whose goals are in opposite directions from the initial position of the agent; then, the well-trained policies for the tasks require contradicting actions.
Thus, in this case, the performance is very poor before the parameters are updated.

Another type of meta-learning methods is called {\it context-based meta-learning}.
PEARL~\cite{rakelly2019efficient} is a state-of-the-art meta-RL method that belongs to this type.
PEARL learns how to infer task information in meta-training and uses it in meta-testing.
Because of this inference, PEARL generally needs less data to improve its performance in meta-testing than methods that update the parameters of neural networks when the tasks in meta-training and meta-testing are similar. 
In addition, the policy and Q-function in PEARL are trained in an off-policy manner, which can further improve its sample efficiency.   
Indeed, PEARL was shown to be more sample efficient than MAML~\cite{rakelly2019efficient}.  

In this paper, we extend the idea of PEARL and propose a novel meta-RL method, \textit{embedding learning and evaluation of uncertainty} (ELUE), which has the following features:
\begin{description}
\item[Off-policy embedding training]~\\
  In PEARL, policy training is based on an off-policy method, but the training for task embedding, which is used for calculating distributions over tasks, depends on what the current policy is, i.e., it is on-policy.
  By removing the dependency on the current policy from the task embedding learning, we propose a fully off-policy method.
  Thanks to policy-independent embedding, the training objective is expected to be stable, and the data collected by past policies can be reused.
\item[Policy and Q-function conditioned by beliefs over tasks]
  PEARL introduces distributions over tasks, i.e. beliefs, but both its policy and Q-function depend on a task variable sampled from the distribution.
  After the task variable is sampled, the variable contains no information on the uncertainty over the tasks.  
  By contrast, ELUE conditions the policy and Q-function on the belief over tasks, which can be used to evaluate uncertainty.
  The use of such policy and Q-function has been shown lead to more precise exploration~\cite{humplik2019meta, zintgraf2020varibad}. 
  In addition, for learning these functions, we apply the information bottleneck objective~\cite{tishby1999information} for generalization, instead of naively maximizing the cumulative reward.

\item[Combination of belief and parameter update]~\\
  PEARL does not update the parameters of the policy and Q-function in meta-testing.
  When there are large differences between the tasks in meta-training and in meta-testing,
  the pretrained belief model, policy, and Q-function may no longer be useful, and PEARL can fail to improve the performance.
  To alleviate this drawback, our method performs not only inference but also parameter update.
\end{description}
%
%
%
We compare the sample efficiency of state-of-the-art meta-RL methods and ELUE through experiments in the Meta-World~\cite{yu2019meta} environment and show that ELUE performs better than those methods.

\section{Preliminaries}
Markov decision processes (MDPs) are models for reinforcement learning (RL) tasks.
An MDP is defined as a tuple $(\mathcal{S}, \mathcal{A}, T, R, \rho)$,
where $\mathcal{S}$ and $\mathcal{A}$ are the state and the action spaces respectively,
$R: \mathcal{S} \times \mathcal{A} \times \mathbb{R} \rightarrow [0,1]$ is a reward function 
that determines the probability of the reward amount and $T: \mathcal{S} \times \mathcal{A} \times \mathcal{S} \rightarrow [0,1]$ is a transition function that determines the probability of the next state.
$\rho$ is the initial state distribution.
Let us denote the policy as $\pi$, which determines the probability of choosing an action at each state.
The objective of RL is to maximize the expected cumulative reward by changing the 

We assume that each task in meta-training and meta-testing can be represented by an MDP, where $\mathcal{S}$ and $\mathcal{A}$ are the same among tasks.
In addition, we consider tasks to be the same when they differ only in $\rho$ because the difference in $\rho$ does not change the optimal policy.
Thus, a different task means different $T$ or $R$.

In our problem, it is assumed that the reward and the transition function are not observable directly.   
We treat this problem as a partially observable MDP (POMDP) and introduce a probability over $R$ and $T$, which is called a {\it belief} \cite{humplik2019meta, zintgraf2020varibad}.
For clarity, let us assume that $R$ and $T$ are parameterized by $\varphi$ and denote them as $R_\varphi$ and $T_\varphi$.  
It is known that a POMDP can be transformed into a belief MDP whose states are based on beliefs and that the optimal policy of the belief MDP is also optimal in the original POMDP~\cite{kaelbling1998planning}.
We denote a history as $h_t := (s_0, a_0, r_0, s_1, \dots, s_t)$, where $s_\tau \in \mathcal{S}$, $a_\tau \in \mathcal{A}$, $r_\tau \in \mathbb{R}$ are the state, the action, and the reward at time $\tau$, respectively. 
In our problem, a belief at time $t$ is $\tilde{b}_t(\varphi) := P(\varphi | h_t)$, and the state of the belief MDP at time $t$ is $s^+_t = (s_t, \tilde{b}_t)$, which is often called a hyper-state. 
The objective of our problem is maximizing $\E_{h_\infty}[\sum_{t=0}^{\infty} \gamma^t r_t]$ by changing a policy that is conditioned on a hyper-state, where $\gamma$ is a discounted factor.  
%
In our problem, the belief is updated by observations:
\begin{align}
&\tilde{b}_{t+1}(\varphi) = P(\varphi | h_{t+1}) \nonumber\\
&\propto P(\varphi) \prod_{\tau = 0}^{t}R_\varphi(s_\tau, a_\tau, r_\tau)T_\varphi(s_\tau,a_\tau, s_{\tau+1}) \label{propto:belieftotal}\\
&\propto P(\varphi| h_t)R_\varphi(s_t, a_t, r_t)T_\varphi(s_t,a_t, s_{t+1}). 
\end{align}
However, in general, the exact calculation of this belief update is intractable, so the existing methods approximate beliefs and avoid the direct calculation~\cite{humplik2019meta,zintgraf2020varibad,igl2018deep,kapturowski2018recurrent}.
In the next section, we introduce the approximated belief model, its update, and other parts of our method.

\section{Method}
In this section, we introduce our method, \textit{embedding learning and evaluation of uncertainty} (ELUE), which learns how to infer tasks and how to use beliefs based on embeddings of task features.
In addition, to alleviate the gap between meta-training and testing, which may prevent improvements from being made if only belief updating is done,
it also adapts the learned policy and Q-function to the meta-test task through the updating of their parameters. 
\begin{figure}[t]
  \centering
   \includegraphics[width=0.98\hsize]{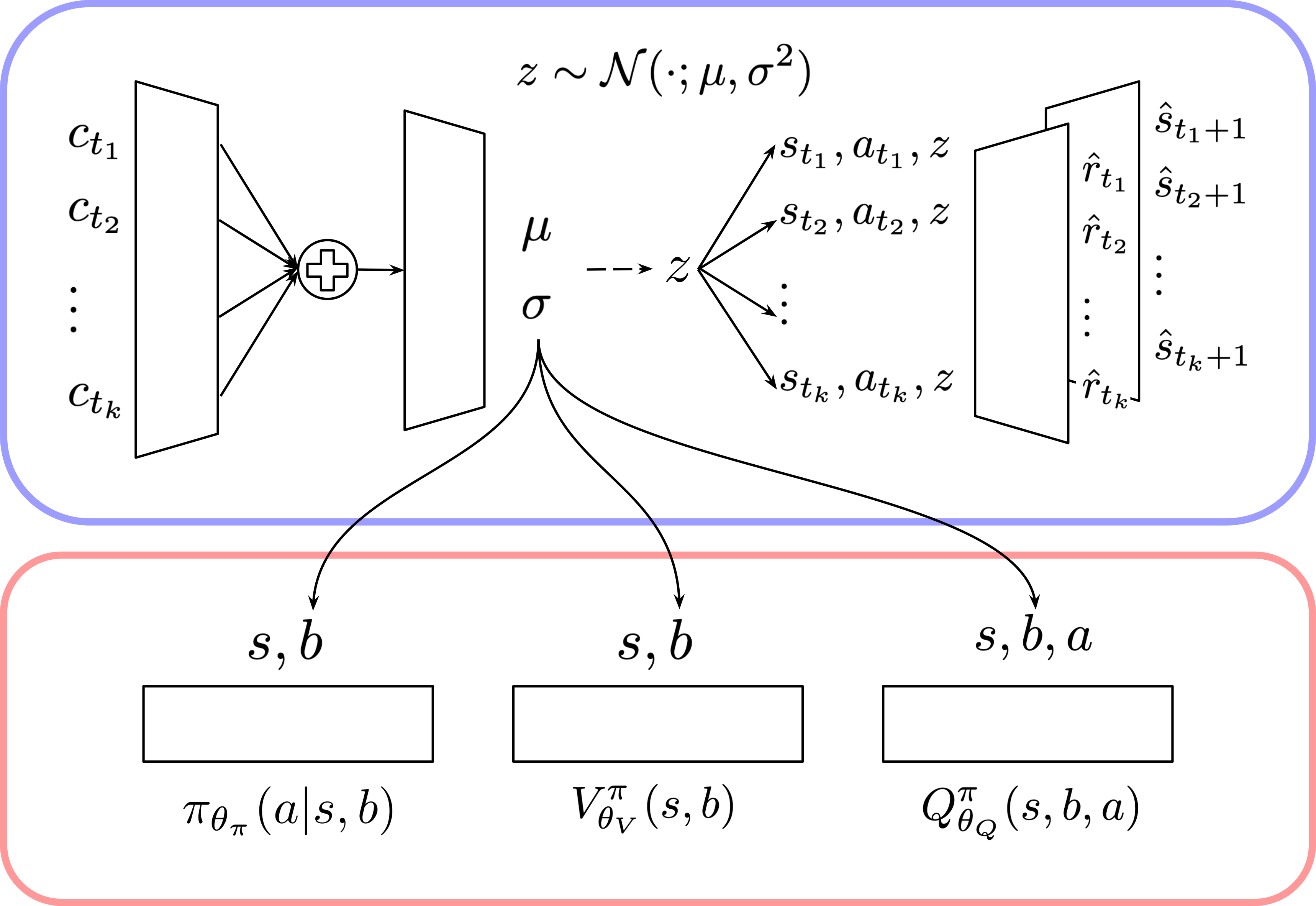}
   \caption[]{A sketch of the architecture of networks in ELUE.
     The networks in the blue box are for the embedding, consisting of an encoder (left) and decoders (right).
     The input of the encoder is a set of $c_t = (s_t, a_t, r_t, s_{t+1})$ and it outputs the parameters of a Gaussian distribution, as introduced in Equation~\eqref{eq:deepset}.
     A part of the inputs of the decoders, $z$, is sampled from the Gaussian distribution, and the decoders predict rewards and next states.  
     In the red box, there are networks for the policy, V-function, and Q-function, which are trained in a similar way to soft actor-critic~\cite{haarnoja2018soft}.
     These networks are conditioned on a belief and use the outputs of the encoder as the belief.  
     Only the networks in the red box are adapted in meta-testing, while all of the networks are pretrained in meta-training. 
   }
  \label{fig:sketch}
\end{figure}
A sketch of the architecture of our networks is shown in \Figref{fig:sketch}.

\subsection{Learning Embedding}\label{sec:embedding}
In meta-training, ELUE learns embeddings for task features.
In this section, we introduce its theoretical background.

We formulate the embedding learning problem as follows.
There is a latent task variable $z$, whose density is $p(z)$,  
and let us assume that the reward $r_t$ and next state $s_{t+1}$ are sampled from a parameterized model $p_\phi(r_t, s_{t+1}|s_t, a_t,z)$, which is shared across tasks.
If $h_t$ is observed frequently, a reasonable model is expected to give $h_t$ a high density.
Thus, in proportion to the frequency of $h_t$, maximizing the density,
$\log \int p(s_0)\prod_{\tau = 0}^{t-1} p_\phi(r_\tau, s_{\tau + 1} | s_\tau, a_\tau, z)\pi(a_\tau | h_\tau)p(z)dz$
leads to a reasonable model.
However, this objective depends on the initial distribution and the current policy.
In terms of belief estimation, they do not contribute to the belief as shown in proportional expression \eqref{propto:belieftotal}.
Thus, instead of the density, we consider the maximization of the ELBO of the following value.
\begin{align}
\log  \int \prod_{\tau = 0}^{t-1} p_\phi(r_\tau, s_{\tau + 1} | s_\tau, a_\tau, z)p(z) dz
\end{align}
%
We introduce a parameterized variational distribution $q_\phi$, and the ELBO is:
\begin{align}
&\log \int \prod_{\tau = 0}^{t-1}p_\phi(r_\tau, s_{\tau + 1}|s_\tau, a_\tau, z)p(z) dz\\
&\geq  \E_{q_\phi(z|h_t)}[\sum_\tau \log p_\phi(r_\tau, s_{\tau + 1} |s_\tau, a_\tau, z)]\nonumber \\
  &- D_{KL}(q_\phi(z|h_t)||p(z)).
\end{align}
We maximize this ELBO in a similar way to a conditional variational autoencoder~\cite{sohn2015learning}, i.e., optimizing the parameters of encoder $q$ and decoder $p$. 
The sum of log-likelihoods in the ELBO is permutation-invariant in terms of time $t$ of tuple $c_t := (s_t, a_t, r_t, s_{t+1})$.
We introduce the following structure so that $q_\phi(z|h_t)$ is also permutation-invariant, because the order of tuples can be ignored to estimate the belief, as shown in expression \eqref{propto:belieftotal}.

As shown in \citet{zaheer2017deep}, a function $q(X)$ is invariant to the permutation of instances in $X$, iff it can be decomposed into the form $g (\sum_{x \in X} f(x))$.
We follow this fact and, instead of history-conditional posterior $q_\phi(z|h_t)$, we use a posterior conditioned on a set of tuples,
\begin{equation}q_{\phi}(z|c_{0:t-1}) := \mathcal{N}\left(z ; g_{\phi}\left(\sum_{\tau = 0}^{t-1} f_{\phi}(c_\tau)\right)\right),\label{eq:deepset}\end{equation} where $\mathcal{N}(\cdot)$ is a Gaussian distribution, and $g_{\phi}(\sum_{\tau = 0}^{t-1} f_{\phi}(c_\tau))$ outputs the parameters of the distribution.
Note that $q_{\phi}(z|c_{0:t-1})$ can be used as an approximated belief over $z$, i.e., $b_t(z)$ and that it can be updated with low computational cost.

Let us denote a replay buffer of tuples of task $i$ as $D_i$ and a set of sampled tuples $c_{t_1}^i, c_{t_2}^i, \dots c_{t_k}^i$ from $D_i$ as $c^i_{t_{1:k}}$.
We define the loss of embedding, $\mathcal{L}_{embed}(\phi)$, as
\begin{align}
&\E_{i,c^i_{t_{1:k}}}[ -\E_{q_{\phi}(z|c^i_{t_{1:k}})}[\sum_{c_\tau \in c^i_{t_{1:k}}} \log p_{\phi}(r_\tau, s_{\tau + 1} | s_\tau, a_\tau, z)]\nonumber\\
&  + D_{KL}(q_{\phi}(z|c^i_{t_{1:k}})||p(z)) ].\label{embed_loss}
\end{align}
Note that this loss function does not depend on the policy.
Thus, for the embedding training, we can reuse the data in the replay buffer collected by past policies, the amount of which is generally large.

This can contribute to the stability of the training objective.
In our implementation,  we use two decoders whose outputs are the probability of reward, $p_\phi(r_t|s_t,a_t,s_{t+1},z)$, and that of next state, $p_\phi(s_{t+1}|s_t,a_t,z)$.

\subsection{Learning Belief-Conditional Policy and Q-Function}\label{sec:learning_policy_and_q}
The ELUE agent learns a belief-conditional policy and Q-function in meta-training.
Instead of simply maximizing the cumulative reward by changing its belief-conditional policy,
the agent maximizes it with an additional information bottleneck (IB) objective for generalization~\cite{tishby1999information}.
IB is a kind of regularization based on mutual information.  
While conventional deep RL methods face the problem of generalization~\cite{zhang2018study,zhao2019investigating,cobbe2019quantifying},
it has been shown that this problem can be alleviated by applying IB~\cite{goyal2018infobot,igl2019generalization}.

We applied the IB loss based on InfoBot~\cite{goyal2018infobot}, which is an application of IB for reinforcement learning.
In their method, it is assumed that goal information is given explicitly for each task.
In our case, goal information is not given, so we use beliefs instead.
We introduce policy $\pi$ which can be decomposed into the form $\pi(a_t | s_t^+) = \int \pi^1(w_t| s_t^+) \pi^2(a_t|w_t, s_t) dz_t$,
where $w_t$ is an additional variable for the IB objective.
In addition, we add mutual information $I(w_t; \tilde{b}_t | s_t)$ as a penalty to the cumulative reward, and the objective is
\begin{align}
  &\E\left[\sum_{t=0}^{\infty} \gamma^t r_t \right] - \beta \sum_{t=0}^{\infty} \gamma^t I(w_t;\tilde{b}_t | s_t) \\ 
 =&\E\left[\sum_{t=0}^{\infty} \gamma^t\left\{ r_t  - \beta \E_{w_t}\left[\log \frac{p(w_t| s_t^+)}{p(w_t| s_t)}\right]\right\}\right] \label{eq:original_objective_2}
\end{align}
The equation is derived from $I(w_t; \tilde{b}_t | s_t) = \E[\log \frac{p(w_t, \tilde{b}_t| s_t)}{p(w_t| s_t)p(\tilde{b}_t| s_t)}]= \E[\log \frac{p(w_t| s_t^+)}{p(w_t| s_t)}]$.
This mutual information penalty is expected to help obtain a task-independent yet useful representation as much as possible.

However, $p(w_t | s_t)$ in Equation~\eqref{eq:original_objective_2} is difficult to compute because of the marginalization by $\tilde{b}_t$ (by $h_t$).
We thus approximate it by a variational distribution $q$ as in previous work~\cite{alemi2016deep, goyal2018infobot, igl2019generalization}.
For any random variables $X,Y,Z$, $I(X;Y|Z)=\E[\log \frac{p(X|Y,Z)}{p(X|Z)}]$ and KL divergence $D_{KL}(p(X|Z)||q(X|Z)) \geq 0$.
Thus, $I(X;Y|Z)\leq\E[\log \frac{p(X|Y,Z)}{q(X|Z)}]$.
By using the fact, a lower bound of \eqref{eq:original_objective_2} is derived and it is
\begin{align}
  \E\left[\sum_{t=0}^{\infty} \gamma^t\left\{ r_t  - \beta \E_{w_t}\left[\log \frac{p(w_t| s_t^+)}{q(w_t| s_t)}\right]\right\}\right].
\end{align}
Although any $q$ is allowed and \citet{alemi2016deep} used a Gaussian as $q$, for simplicity, we fix $q(w_t| s_t^+)$ to be a uniform distribution.
By removing a constant part, the objective is 
\begin{align}
 \E\left[\sum_{t=0}^{\infty} \gamma^t \{r_t  - \beta \E_{w_t}[\log p(w_t| s_t^+) ]\} \right],
\end{align}
and ELUE maximizes this value.

This objective is a variant of the objective of soft actor-critic (SAC)~\cite{haarnoja2018soft}, which is one of the most sample efficient off-policy RL methods.  
We follow the same way as SAC and update the policy, Q-function, and V-function.
Let us denote a belief conditioned on the tuple set $c^i_{t_{1:k-1}}$ as $b^i$ and the belief updated from $b^i$ by an additional tuple, $c^i_{t_{k}}$, as $b'^i$.
For simplicity, we abbreviate the subscript $t_{k}$ in $c^i_{t_{k}}$ and denote it as $(s^i,a^i,r^i,s'^i)$.
ELUE minimizes $\mathcal{L}_{actor}(\theta_\pi)$, $\mathcal{L}_{critic}^Q(\theta_Q)$, and $\mathcal{L}_{critic}^V(\theta_V)$, which are respectively,
\begin{align}
  &\E_{i,c^i_{t_{1:k}}}[\E_{w, a}[\beta \log \pi_{\theta_\pi}(w|s^i,b^i) - Q_{\theta_Q}(s^i,b^i,a)]], \label{actor_loss}\\
  &\E_{i,c^i_{t_{1:k}}}[(Q_{\theta_Q}(s^i, b^i, a^i) - \hat{Q}(s^i,b^i,a^i))^2], \;\mbox{and} \label{Q_critic_loss}\\
  &\E_{i,c^i_{t_{1:k}}}\left[(V_{\theta_V}(s^i) - \hat{V}(s^i))^{2}\right], \;\mbox{where} \label{V_critic_loss}\\
&\hat{Q}(s^i, b^i, {a}^i)= r^i + \gamma V_{\bar{\theta}_V}\left({s}'^i, b'^i \right),\label{Q_hat}\\
&\hat{V}(s^i) = \E_{w, a}[Q_{\theta_Q}(s^i,b^i,a) -\beta \log \pi_{\theta_\pi}(w|s^i,b^i)]\label{V_hat} 
\end{align}
and $\bar{\theta}_V $ is a parameter vector that is updated by $\bar{\theta}_V \leftarrow (1- \lambda) \bar{\theta}_V + \lambda\theta_V$.
We show the procedures of our method in meta-training in \Algref{alg:meta-training}.
Our belief is based on the random sampling of tuples.
Because of that, this training depends less on actual trajectories than training with naive trajectory-based beliefs.  
This can be an advantage, as PEARL with a naive trajectory-based encoder was not so effective as that with an encoder based on sampled tuples~\cite{rakelly2019efficient}.

In addition to the policy, we also apply IB to the Q-function and V-function.
To stablize the target values i.e. $\pi$ and $Q$ in \eqref{V_hat}, $V$ in \eqref{Q_hat}, and $Q$ in \eqref{actor_loss}, 
we use the mean values of the additional variables for IB for the V-function, Q-function, and policy, instead of sampled variables in the same way as \citet{igl2019generalization}.
This is expected to reduce fluctuations by introducing IB. 

Here, we describe more details on the implementation of our method.
First, to reduce the computational cost, instead of naively allocating one random sampled tuple set $c^i_{t_{1:k-1}}$ and belief conditioned on it $b^i$ to one additional tuple $c^i_{t_k}$, we allocate the same tuple set to some additional tuples.
Second, to train the agent in a variety of situations in terms of the amount of data necessary to infer a task, we randomly sample $k$, the number of tuples in $c^i_{t_{1:k-1}}$.
Third, for $\pi^1(w_t | s_t^+)$ and $\pi^2(a_t | w_t, s_t)$, we use Gaussian distributions, and for bounding actions, we use tanh as in SAC~\cite{haarnoja2018soft}.
\begin{algorithm}[tbh]
\caption{Meta-training}
\label{alg:meta-training}
\begin{algorithmic}[1] 
  \STATE A set of meta-training tasks, $\mathcal{T}$ is given 
\WHILE{not done}
\STATE Sample tasks from $\mathcal{T}$
\STATE Initialize beliefs 
\FOR{$i \in$ the sampled tasks}
\FOR{step in data collection steps}
\STATE Gather data from task $i$ by policy $\pi(\cdot | s^i, b^i)$
\STATE Update belief $b^i$ and replay buffer $D^i$    
\ENDFOR
\ENDFOR
\FOR{step in training steps}
\STATE Make a set of tasks $\mathcal{T}'$ randomly from $\mathcal{T}$ 
\STATE Calculate $\mathcal{L}_{embed}$ for $\mathcal{T}$, as in formula \eqref{embed_loss}, and update parameters to minimize $\mathcal{L}_{embed}$
\STATE Calculate $\mathcal{L}_{actor}$ and $\mathcal{L}_{critic}$ for $\mathcal{T}'$, as in formulae \eqref{actor_loss}, \eqref{Q_critic_loss}, and \eqref{V_critic_loss}, and update parameters to minimize $\mathcal{L}_{actor}$ and $\mathcal{L}_{critic}$ 
\ENDFOR
\ENDWHILE
\end{algorithmic}
\end{algorithm}

\subsection{Adaptation in Meta-Test}\label{sec:meta-testing}
\begin{algorithm}[tb]
\caption{Meta-testing}
\label{alg:meta-testing}
\begin{algorithmic}[1] 
\STATE A meta-test task is given
\WHILE{not done}
\FOR{step in data collection steps}
\STATE Gather data from meta-test task by policy $\pi(a | s, b)$
\STATE Update belief $b$ and replay buffer $D$ 
\ENDFOR
\FOR {step in training steps}
\STATE Calculate $\mathcal{L}_{actor}$ and $\mathcal{L}_{critic}$ for the task, as in formulae \eqref{actor_loss}, \eqref{Q_critic_loss}, and \eqref{V_critic_loss} and update parameter to minimize modified $\mathcal{L}_{actor}$ and $\mathcal{L}_{critic}$
\ENDFOR
\ENDWHILE
\end{algorithmic}
\end{algorithm}
In meta-testing, the ELUE agent collects data based on the policy conditioned on a belief.
The belief is updated at every time step.
After updating the belief enough times, our method updates the parameters of neural networks to alleviate the gap between tasks in meta-training and meta-testing.
Pseudo code is shown in \Algref{alg:meta-testing}.
In meta-testing, there are differences in the parameter update in meta-training:

1) The parameters for embedding, $\phi$, are fixed to avoid catastrophic forgetting~\cite{french1999catastrophic} about what it learned in meta-training.
Naive updating of the parameters about embedding in meta-testing leads to catastrophic forgetting because the number of tasks in meta-testing is one, which means that the decoder can reconstruct the reward and next state without the latent task variable information, if the decoder is sufficiently trained in meta-testing.
If the output of the decoder is independent from the latent task variable, only the second term of \eqref{embed_loss} is relevant to the learning of the encoder, which means that the encoder loss is minimized when its output is the same as that of the prior, $p(z)$.
On the basis of these considerations, we fix $\phi$ in meta-testing.

2) After updating belief enough times, we fix and copy it for each neural network, and train it with the parameters of each network by gradient descent.
This expected to help adaptation to the meta-test task by giving additional flexibility to task information processing.
This update method is inspired by semi-amortized variational autoencoder~\cite{kim2018semi}, whose update method is a combination of inference and gradient descent.
  By combining gradient descent, it showed better asymptotic performance. 


\comm{
\memo{１．について，計算量の削減できるという良さがある．updateしても意味が無いと予想される，実際に，素早く改善するので，updateするまでも無い（でも改善する可能性があるのは否定できない）１．について書くことは全くもって無意味でも無い．．．　　　２．について} 
結局fixするので，．．．fixに至る過程，選択肢の限定としての価値があるので，とくに悩むことは無い．
embedを学習するなら，fixという選択肢は無い
ELUE fixes the belief after updating it enough times, empricallyにはそれで十分というか特に
to stabilize because it depends on which tuples are used for calculation, which can be noise. 
\memo{contextの数を固定しているから，FIFOの形で処理している旨述べて，．．．とりあえず保留}
}

\begin{figure*}[h!]
  \begin{subfigmatrix}{3}
 \subfigure[basket-ball]{
   \includegraphics[width=0.32\hsize]{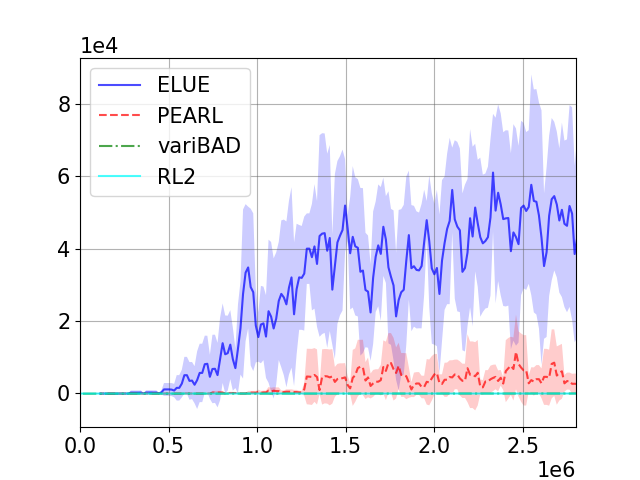}}
 \subfigure[dial-turn]{
   \includegraphics[width=0.32\hsize]{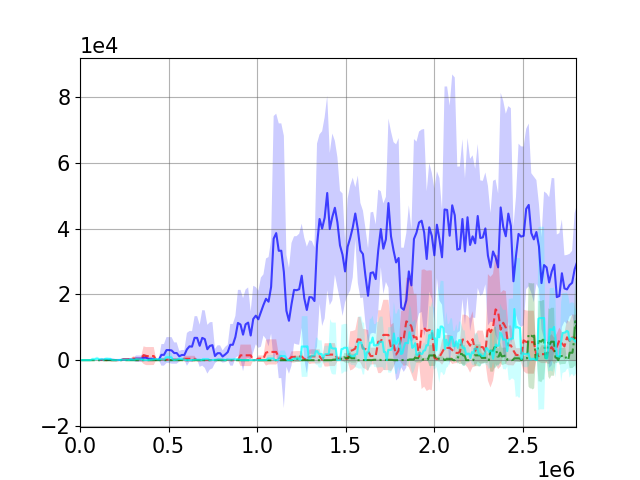}}
 \subfigure[pick-place]{
   \includegraphics[width=0.32\hsize]{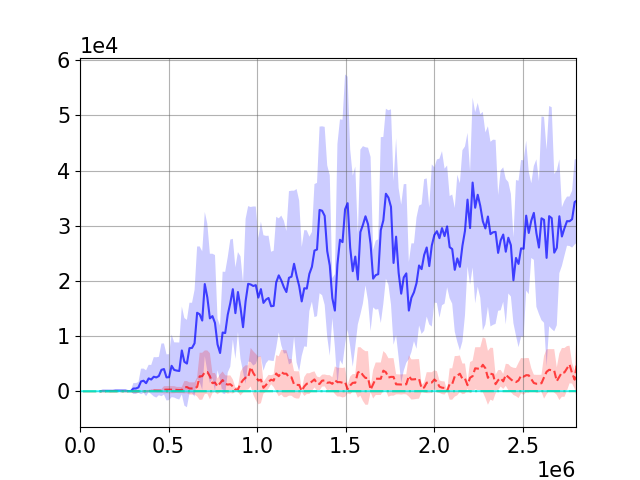}}
 \subfigure[reach]{
   \includegraphics[width=0.32\hsize]{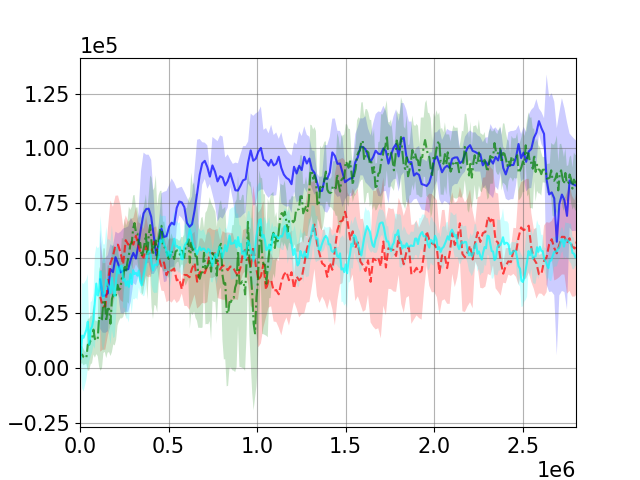}}
 \subfigure[sweep-into]{
   \includegraphics[width=0.32\hsize]{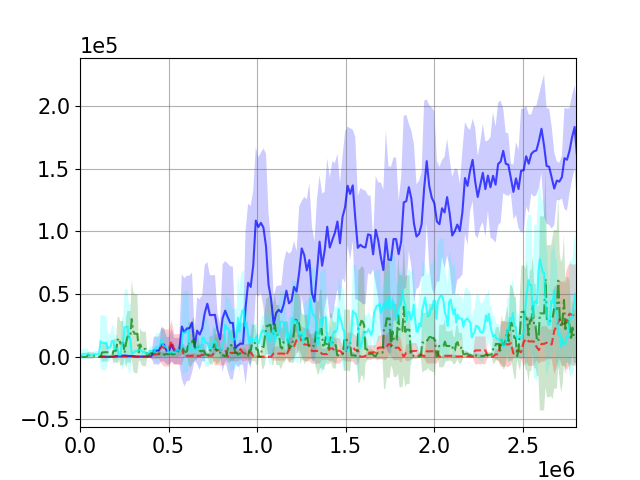}}
 \subfigure[window-open]{
   \includegraphics[width=0.32\hsize]{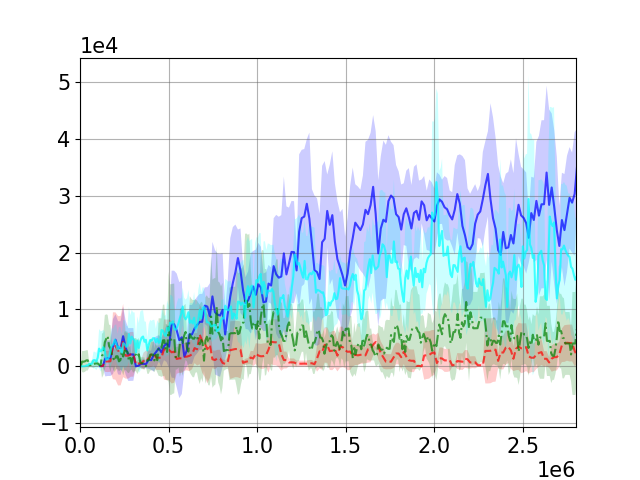}}
 \end{subfigmatrix}
\caption[]{The comparison of sample efficiency of each method in ML1 benchmarks.  The vertical axis is the average rewards in the first episode in meta-testing and the horizontal axis is the number of time steps in meta-training.}
  \label{fig:validation0_ml1}
\end{figure*}

\section{Related Work}\label{relatedworks}
In this section, we review existing methods related to our method and discuss the differences between them.

Our method is inspired by PEARL, but there are essential differences.
First, PEARL has no decoder, and the encoder is trained to minimize the critic loss.
It is a simple approach, but its embedding can change depending on the current policy.
The performance of PEARL was shown to degrade when used with off-policy (i.e., not recent) data~\cite{rakelly2019efficient}.
Therefore, PEARL uses an additional buffer for recent data to avoid the degradation; in contrast, our method can train the embedding using old data and does not need an additional buffer.
Second, PEARL uses an encoder that can be represented as
    $q_{\phi}(z|h_t) \propto \prod_{\tau = 0}^{t-1}\mathcal{N}(z;\mu_\tau, \sigma_\tau^2) \propto \mathcal{N}(z; \frac{\sum_{\tau=0}^{t-1}\frac{\mu_\tau}{\sigma_\tau^2}}{\sum_{\tau=0}^{t-1}\frac{1}{\sigma_\tau^2}},\frac{1}{\sum_{\tau=0}^{t-1}\frac{1}{\sigma_\tau^2}})$,
    where $\mu_\tau$ and $\sigma_\tau^2$, the mean and variance of a Gaussian distribution, are the outputs of the neural network, $f_{\phi}(\cdot)$. 
As discussed around Equation~\eqref{eq:deepset}, this is not a general form for encoder representation in terms of permutation invariance among $c_\tau$, while our encoder is represented in a general form. 
Third, PEARL's policy and Q-function, $\pi(s, z)$ and $Q(s,a,z)$, where $z$ is sampled from $q(z|h_t)$, are $z$-conditional, and $z$ itself has no uncertainty information.
In comparison, ours are belief-conditional, which has uncertainty information.
Fourth, PEARL only considers inference in meta-testing, while our method considers the updating of the parameters of neural networks.

There are other meta-learning methods based on inference; however, these methods are on-policy~\cite{zintgraf2020varibad,humplik2019meta, vuorio2019multimodal, duan2016rl}.
Among these methods, VariBAD~\cite{zintgraf2020varibad} is the most related to our method.
It considers embedding just like ours and beliefs over the embedding space.
However, its sample efficiency is not as high as PEARL as shown in \citet{zintgraf2020varibad}.  
In addition, its encoder is based on recurrent neural networks whose input is simply a history.
Moreover, it does not consider updating the parameters of neural networks in meta-testing.

As for off-policy approaches, guided meta policy search~\cite{mendonca2019guided} is introduced as an off-policy meta-learning method.
Their method is based on MAML.
In meta-training, it updates the parameters to imitate expert trajectories (outer updates) after updates by policy gradients (inner updates), which means that the inner updates and the updates in meta-testing need on-policy data.
In addition, it is not based on inference.
\citet{lin2020model} proposed an off-policy method which uses a decoder and policy conditioned the parameters of the decoder. The parameters corresponds to a belief in our method and they are adapted to a meta-test task by gradient descent.
Their method is also not inference.
It is based on MAML-like parameter updates.

\section{Experiments}\label{experiments}
In this section, we compare the sample efficiency of the existing ``context-based'' methods (PEARL, variBAD, and RL2~\cite{duan2016rl}) and our method.
In the conventional setting of meta-learning experiments, the tasks in meta-testing and meta-training were in the same category and the tasks differed only in goal positions or weights of torso of the agent~\cite{finn2017model, rakelly2019efficient, zintgraf2020varibad, vuorio2019multimodal}.
Our experimental settings include not only the conventional ones but also settings where the types of tasks in meta-testing and meta-training are different. 
The environment of the experiments is Meta-World with MuJoCo 2.0.
Meta-World is a collection of Sawyer robot arm tasks, and there are 50 types of tasks and several benchmarks.  
In a task of Meta-World, each episode length is 150.

First, to examine the sample efficiency of our method, we conducted experiments with the conventional settings.
We followed the ML1 benchmark scheme of Meta-World, where the differences in tasks mean differences in goals and the goals are sampled from the same distribution in meta-training and meta-testing.
We chose six types of tasks, basket-ball, dial-turn, pick-place, reach, sweep-into, window-open, that were chosen from the types of tasks of the ML10 benchmark\footnote{The names of the types of tasks in the Meta-World paper are different from those in the Meta-World implementation. We refer to the names in the implementation.} (descriptions of the types of tasks we used are shown in Table 1 in appendices).  
For each run, 20 meta-training tasks were generated.
We executed five runs with different seeds and evaluated the average episode reward at the first episode in meta-testing.
The results are shown in \Figref{fig:validation0_ml1}. 
The results show that ELUE achieved higher performance than the other methods.

\begin{figure}[h!]
   \includegraphics[width=0.9\hsize]{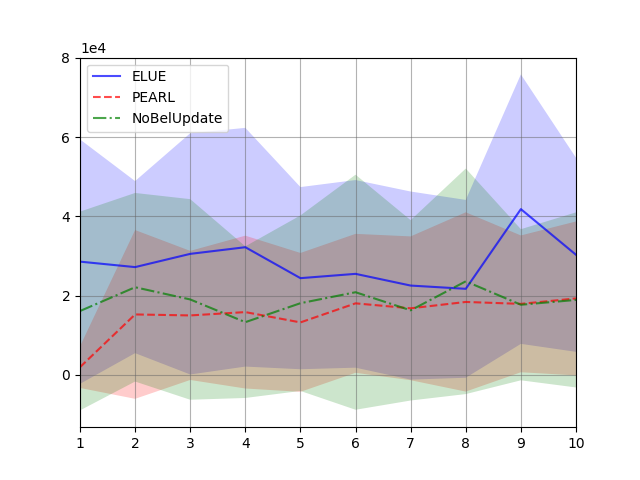}
   \caption[]{The comparison of performance in meta-test of pick-place on ML1 benchmark by inference.  The vertical axis is the average episode reward and the horizontal axis is the number of episodes.}
  \label{fig:inference}
\end{figure}

Second, to see the merits of our belief updates in meta-testing, 
we examined the performance of ELUE and PEARL in meta-testing with only inferences (i.e. belief updates in ELUE, and posterior updates and sampling the task variable in PEARL) and that of ELUE without any update of a belief, i.e. keeping a belief being a prior (``NoBelUpdate'') for comparison.
In the experiment, meta-trained networks at about three million time steps in the first experiment of pick-place were used.
The results are shown in \Figref{fig:inference}.
The results show that both methods only need small amounts of data to improve the performance. Especially, the performance of ELUE was relatively high from the first episode while that of PEARL was low at first and improved in the next episode.

\begin{figure*}[tbh]
  \begin{subfigmatrix}{3}
 \subfigure[The first episode rewards]{
   \includegraphics[width=0.32\hsize]{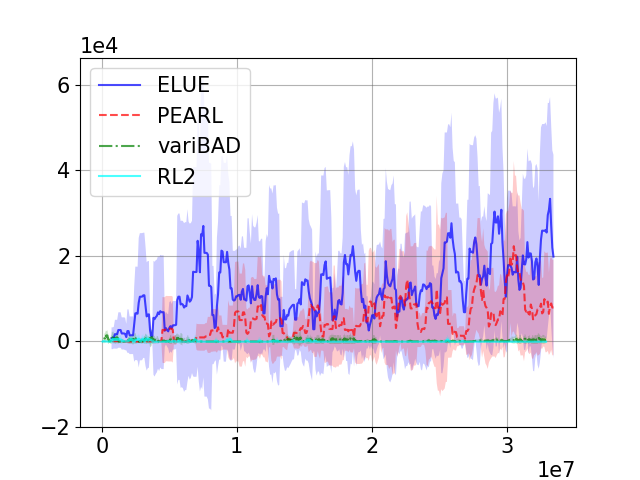}}
 \subfigure[The Third episode rewards]{
   \includegraphics[width=0.32\hsize]{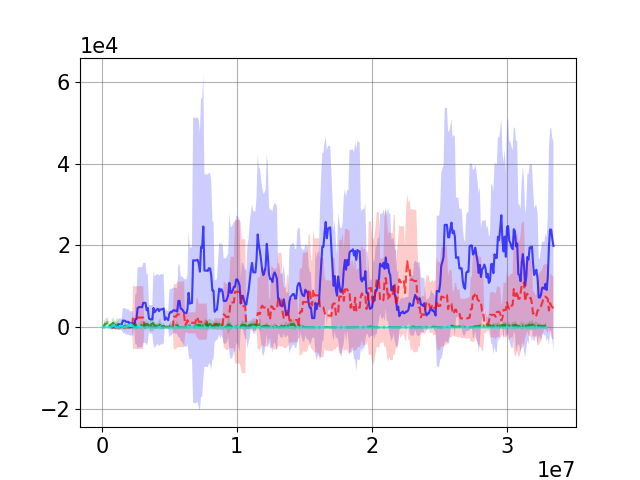}}
 \subfigure[Learning curves]{
   \includegraphics[width=0.32\hsize]{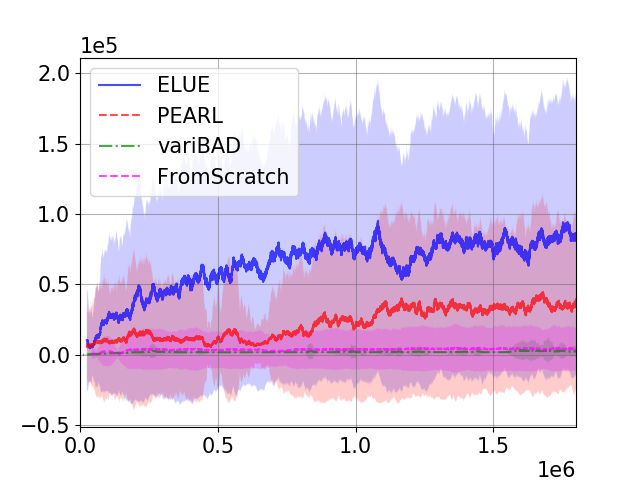}}
 \end{subfigmatrix}
 \caption[]{The comparison of sample efficiency of each method on ML10 benchmark.  The vertical axis is the average episode rewards in meta-testing.  
   The horizontal axis is the number of time steps (a), (b) in meta-training and (c) in meta-testing.}
  \label{fig:ml10}
\end{figure*}

Third, to examine the sample efficiency in a set of diverse tasks, we conducted experiments, following the ML10 benchmark scheme, where the types of tasks in meta-training and meta-testing were different.
In the benchmark, there are 15 types of tasks, ten for meta-training and five for meta-testing.
We evaluated the sample efficiency from two perspectives: the performance in meta-testing in the early stages like~\Figref{fig:validation0_ml1} and their improvement as the amount of data increases.
For an experiment in the latter perspective, we modified PEARL and variBAD to make them update the parameters of neural networks in meta-testing, although the original versions of them do not update the parameters.
In the experiment, meta-trainings were executed five times. For each run, 40 tasks were generated.
For each meta-training, meta-tests were executed six times.
The meta-tests were executed for about two million steps. 
In the meta-tests, we used meta-trained networks after 600 iterations (25 million time steps) for ELUE and PEARL, 4000 iterations (32 million time steps) for variBAD.  
To clarify the amount of improvement with our method, we also executed a variant of ELUE which learns policies from scratch in meta-testing without using beliefs.
%
The results are shown in \Figref{fig:ml10}.
The performances of PEARL and ELUE slightly improved as time steps in meta-training increased.
As for the learning curves of meta-testing, although the variance of the performance of ELUE was large, ELUE outperformed the other methods on average. 

Fourth, to see the effectiveness of applying information bottleneck to our method, we compared the performances of variants of ELUE which was naively combined with soft actor-critic.
We changed the coefficients of entropy in soft-actor critic.
The other hyper parameters were the same among the variants.
  \begin{figure}[tbh]
   \includegraphics[width=\hsize]{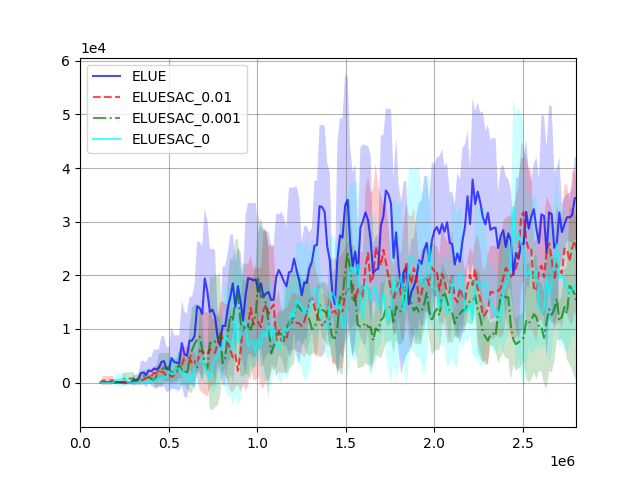}
\caption[]{The comparison of variants of ELUE.  The vertical axis is the average episode rewards in meta-testing and the horizontal axis is the number of time steps in meta-training.}
  \label{fig:IB-SAC}
\end{figure}
The results (\Figref{fig:IB-SAC}) show that ELUE was slightly better than the other variants.
This implies that information bottleneck objectives helped generalization of learned policies as shown in~\citet{igl2019generalization}.

\begin{figure}[tbh]
   \includegraphics[width=\hsize]{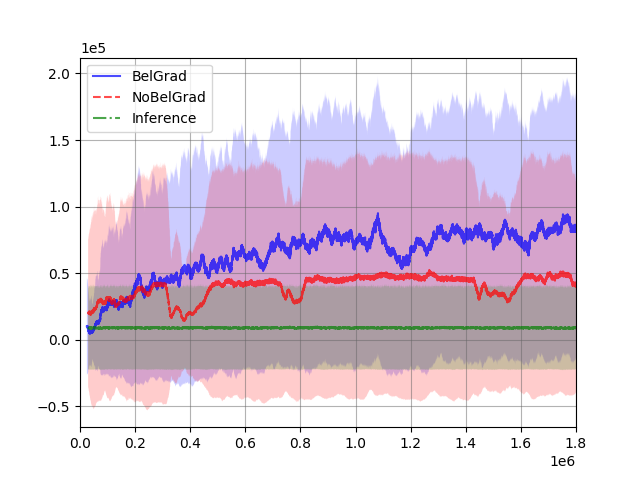}
   \caption[]{The comparison of updating method in meta-testing.  The vertical axis is the average episode rewards in meta-testing and the horizontal axis is the number of time steps in meta-testing.
   ``BelGrad'' is the same as ``ELUE'' in \Figref{fig:ml10}~(c).}
  \label{fig:ml10_meta-test_ab}
\end{figure}

Fifth, to examine the effectiveness of our updating method in meta-testing, we compared the performances of the different update methods (``BelGrad'', ``NoBelGrad'' and ``Inference'') in the ML10 benchmark.
BelGrad means belief updates by gradient descent after inference, which is used in meta-test of ELUE.  
NoBelGrad means that gradient descent updates are applied to not the belief but the parmeters of neural networks.
Inference means no gradient updates.
NoBelGrad and Inference were executed three times for each meta-training.
We used the same settings and the same meta-trained networks in the experiment of \Figref{fig:ml10}.
The results (\Figref{fig:ml10_meta-test_ab}) show that the performance of Inference did not improve as the amount of data increased, and that BelGrad was better than NoBelGrad.

\section{Conclusion}
We have proposed a novel off-policy meta-learning method, ELUE, which learns the embeddings of task features, beliefs over the embedding space, belief-conditional policies and Q-functions.
Because of the belief-conditional policies, the performance can be improved by updating the beliefs, especially when the meta-test task is similar to the meta-training tasks.
ELUE also updates the parameters of neural networks in meta-testing, which can alleviate the gap between tasks in meta-testing and those in meta-training.
In the experiments with Meta-World benchmarks, we examined the sample efficiency of ELUE and existing methods in the two cases where the tasks in meta-training and meta-testing were similar and diverse. The results show that ELUE outperforms the other methods in both cases.
We have also shown merits of the belief-conditional policies and the parameter updates in meta-testing.

\comm{
    \memo{belief-conditional? belief conditional?}
    \memo{a small amount of data}
    \memo{on benchmark}
    \memo{demonstrate <- show ?}
    \memo{the sample efficiency}
    \memo{have shown <- showed ?}
    \memo{``its'' performance}
    \memo{history-conditional}
    \memo{results show that <- results showed that}

}
\bibliography{database}
\clearpage

\section{Appendices}
In this section, we provide supplementary information about Meta-World, ELUE implementation, additional experimental results, and theoretical derivations.
\subsection{Meta-World}
For convenience, we briefly explain Meta-World.
In Meta-World, there are 50 types of tasks using Sawyer robot arm.
The descriptions of 15 types of tasks we use are shown in \Tbref{tb:task_types}.
The dimensions of the observation and action of each task are six and four, respectively.
The amount of reward is determined based on the center positions of fingers of the arm, an object, and a goal.
Here, the goal position cannot be observed by the agent.
In the reach tasks, only the center position of the fingers and the goal is related to the reward. 
In the other 14 types of tasks, the position of the object is also related to the reward, and the agent can get high reward by moving the object to the goal. 

\begin{table*}[hbtp]
  \centering
  \begin{tabular}{ll}
    \hline
    Type of task  & Description  \\
    \hline
    basket-ball   &  Dunk the basketball into the basket. Randomize basketball and basket positions  \\
    button-press-topdown   &  Press a button from the top. Randomize button positions  \\ 
    dial-turn   &    Rotate a dial 180 degrees. Randomize dial positions\\
    drawer-close  &   Push and close a drawer. Randomize the drawer positions \\
    pick-place   &  Pick and place a puck to a goal. Randomize puck and goal positions  \\
    reach   &   Reach a goal position. Randomize the goal positions \\
    sweep   &   Sweep a puck off the table. Randomize puck positions \\
    window-open   &  Push and open a window. Randomize window positions  \\
    push  &   Push the puck to a goal. Randomize puck and goal positions \\
    door  &    Open a door with a revolving joint. Randomize door positions \\ 
    \hline
    door-close   &  Close a door with a revolvinig joint. Randomize door positions  \\
    drawer-open  &   Open a drawer. Randomize drawer positions \\
    lever-pull     &  Pull a lever down 90 degrees. Randomize lever positions  \\
    shelf-place   &   Pick and place a puck onto a shelf. Randomize puck and shelf positions \\
    sweep-into   &  Sweep a puck into a hole. Randomize puck positions  \\
    \hline
  \end{tabular}
  \caption{The description of each type of task in Meta-World, which are extracted from Table 1 in~\cite{yu2019meta}. Above/below types of tasks are used for meta-training/testing in ML10 benchmarks}
  \label{tb:task_types}
\end{table*}

\subsection{Implementation Details}
We provide additional explanations about our implementation.
In our implementation, we execute initial sampling at the beginning of meta-training and meta-testing.
In the sampling phase, the agent collect data (lines 6 -- 9 in \Algref{alg:meta-training}) for all tasks.  
After this phase, we execute initial embedding training.
This is done by a similar way to existing methods for training variational autoencoders~\cite{ha2018world, zintgraf2020varibad}.
In meta-testing, we fix the belief after the initial sampling phase.
This is because we found that the improvement by belief updates was very fast and even one episode was enough for the convergence of the performance, as shown in~\Figref{fig:inference}.


\subsection{ML1 Meta-Testing}
\begin{figure}[tbh]
   \includegraphics[width=\hsize]{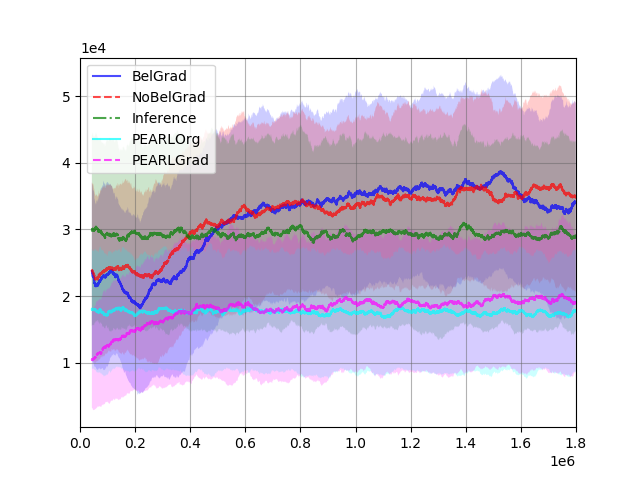}
   \caption[]{The comparison of learning curves in meta-testing.
     ``PEARLOrg'' is original PEARL's update i.e. updating posterior and sampling a task variable from the posterior and ``PEARLGrad'' is PEARL with updating the parameters of neural networks as ``PEARL'' in  \Figref{fig:ml10}~(c).
   The other methods are the same as the methods in \Figref{fig:ml10_meta-test_ab}}
  \label{fig:pick-place_meta-test}
\end{figure}

In addition to the ML10 benchmark (\Figref{fig:ml10_meta-test_ab}), we compare learning methods in meta-testing of pick-place on the ML1 benchmark.
The experimental settings were the same as the experiments of \Figref{fig:inference} except for the number of time steps in meta-testing.
The result is shown in \Figref{fig:pick-place_meta-test}.
The results show that the performances of Inference and PEARLOrg were almost the same from beginning to end, while those of BelGrad, NoBelGrad, and PEARLGrad were improved.
In PEARL and ELUE, the performance of methods with the parameter updates were worse than that of methods without the parameter updates at first but gradually improved, and at last became better.
In \Figref{fig:pick-place_meta-test}, the difference between BelGrad and NoBelGrad was not so clear as that in \Figref{fig:ml10_meta-test_ab}.
\subsection{ML1 Supplemental Results}
\begin{figure}[tbh]
   \includegraphics[width=\hsize]{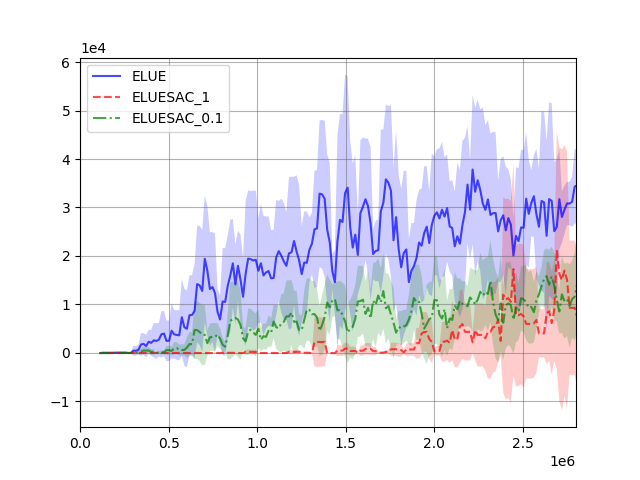}
\caption[]{The comparison of variants of ELUE.  The vertical axis is the average episode rewards in meta-testing and the horizontal axis is the number of episodes.}
  \label{fig:IB-SAC_sup}
\end{figure}

We show supplementary results about \Figref{fig:IB-SAC}.
The results are shown in \Figref{fig:IB-SAC_sup}.
These results are not included in \Figref{fig:IB-SAC} because of its visibility.
The results of \Figref{fig:IB-SAC} and~\ref{fig:IB-SAC_sup} suggest that information bottleneck objectives help generalization of policies.

We also show supplementary results about \Figref{fig:validation0_ml1}.
\begin{figure*}[tbh]
  \begin{subfigmatrix}{3}
 \subfigure[basket-ball]{
   \includegraphics[width=0.32\hsize]{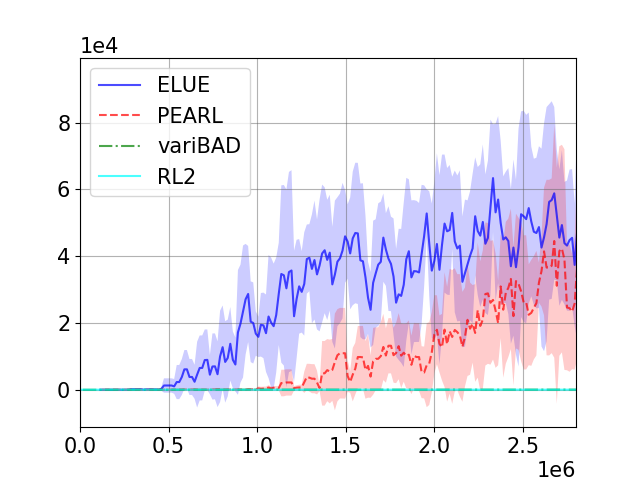}}
 \subfigure[dial-turn]{
   \includegraphics[width=0.32\hsize]{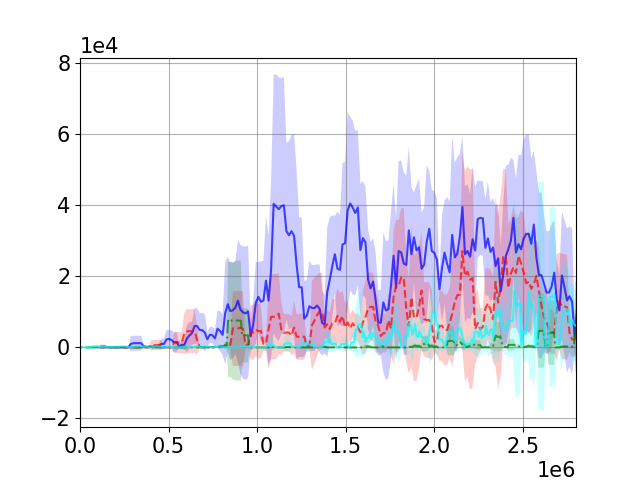}}
 \subfigure[pick-place]{
   \includegraphics[width=0.32\hsize]{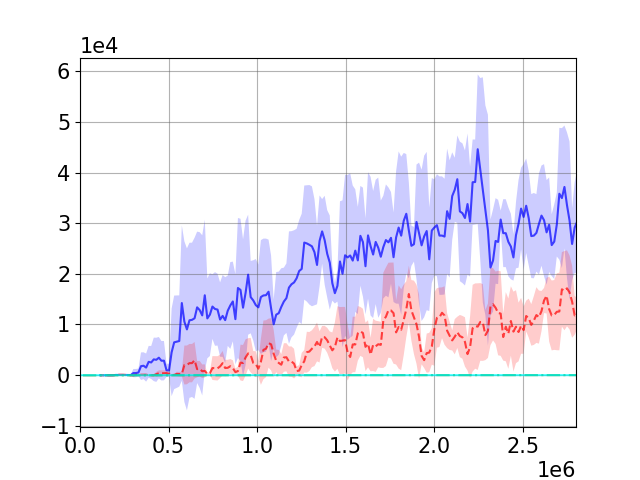}}
 \subfigure[reach]{
   \includegraphics[width=0.32\hsize]{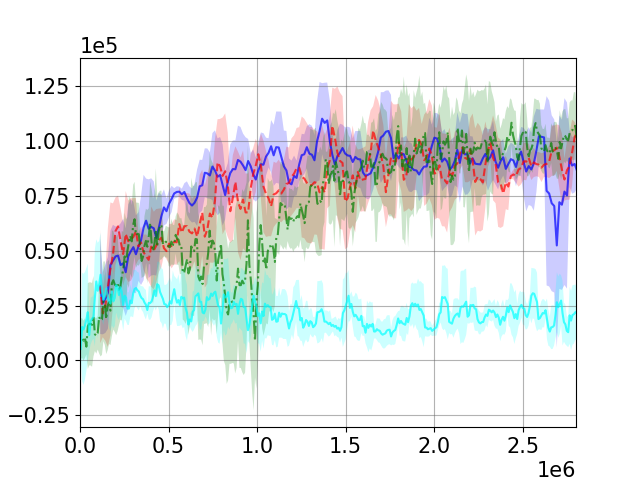}}
 \subfigure[sweep-into]{
   \includegraphics[width=0.32\hsize]{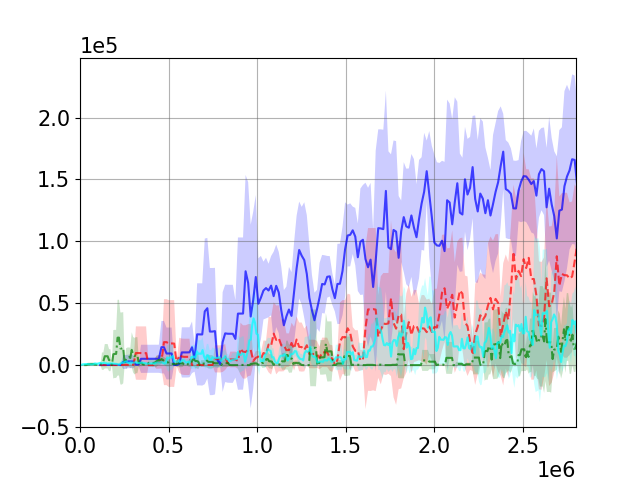}}
 \subfigure[window-open]{
   \includegraphics[width=0.32\hsize]{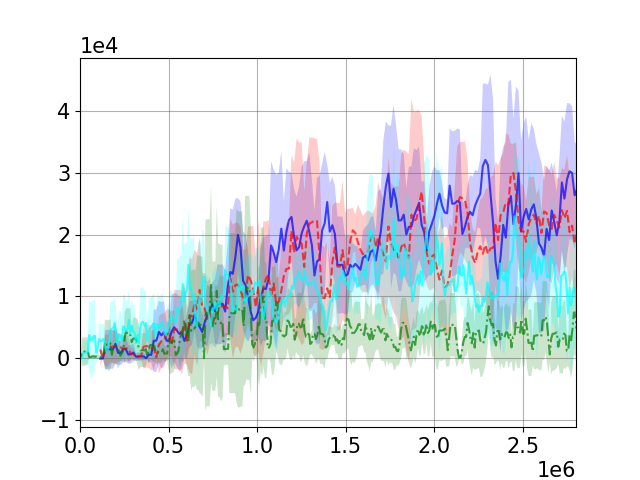}}
 \end{subfigmatrix}
 \caption[]{Comparison of sample efficiency of each method in ML1 benchmarks. The vertical axis is the average rewards in the third episode in meta-testing and the horizontal axis is the number of time steps in meta-training.
 }
  \label{fig:validation2_ml1}
\end{figure*}
\Figref{fig:validation2_ml1} is the third episode rewards of the same experiments of \Figref{fig:validation0_ml1}.
The results show that the performance of PEARL was better than that in the first episode.
However, even in the third episode, ELUE outperformed PEARL in some tasks.

\begin{figure*}[tbh]
  \begin{subfigmatrix}{3}
 \subfigure[First episodes of PEARL]{
   \includegraphics[width=0.32\hsize]{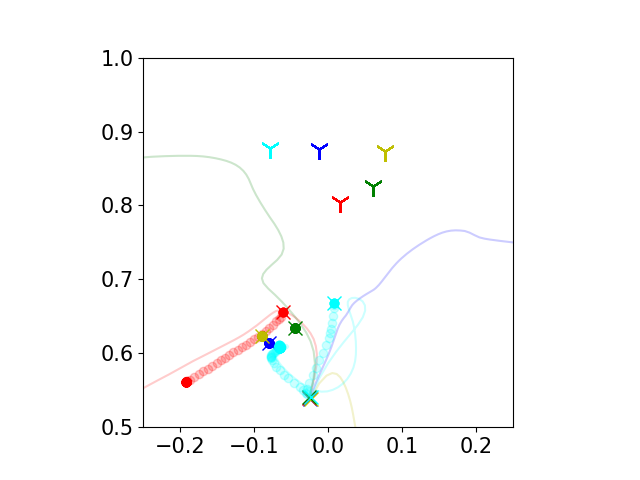}}
 \subfigure[Third episodes of PEARL]{
   \includegraphics[width=0.32\hsize]{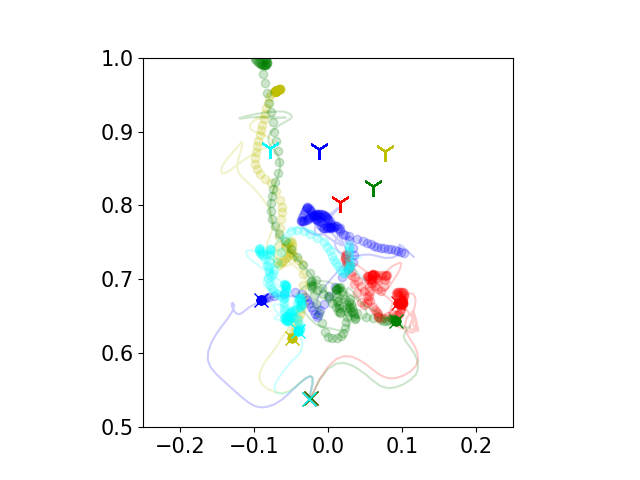}}
 \subfigure[Fifth episodes of PEARL]{
   \includegraphics[width=0.32\hsize]{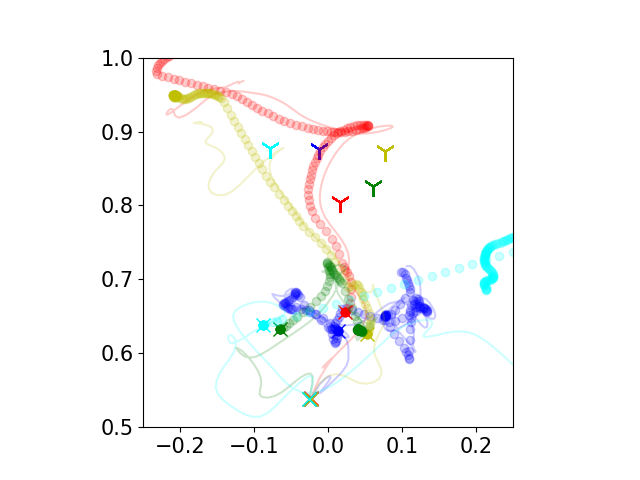}}
 \subfigure[First episodes of ELUE]{
   \includegraphics[width=0.32\hsize]{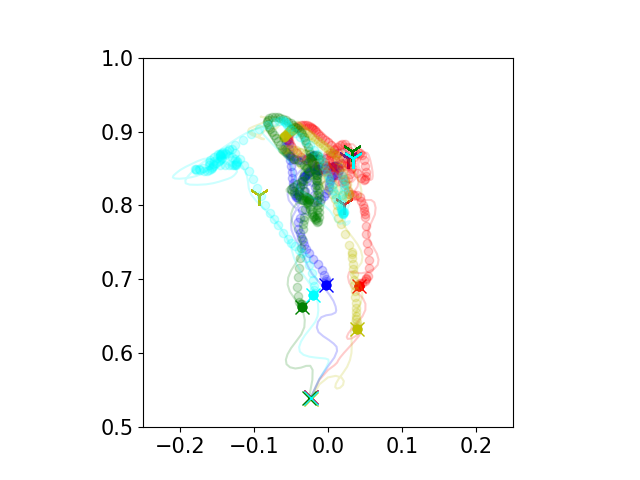}}
 \subfigure[Third episodes of ELUE]{
   \includegraphics[width=0.32\hsize]{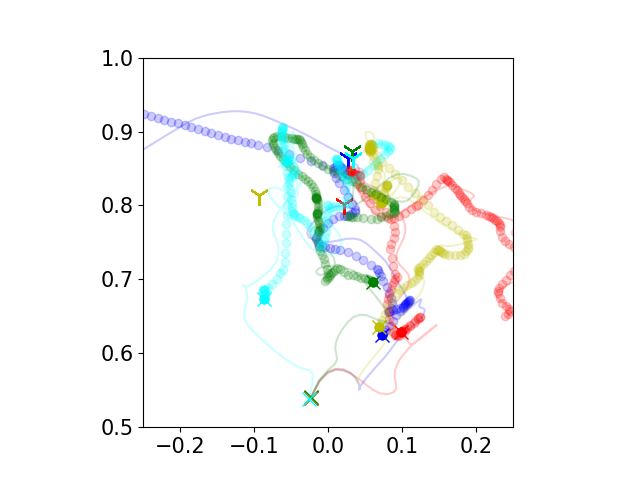}}
 \subfigure[Fifth episodes of ELUE]{
   \includegraphics[width=0.32\hsize]{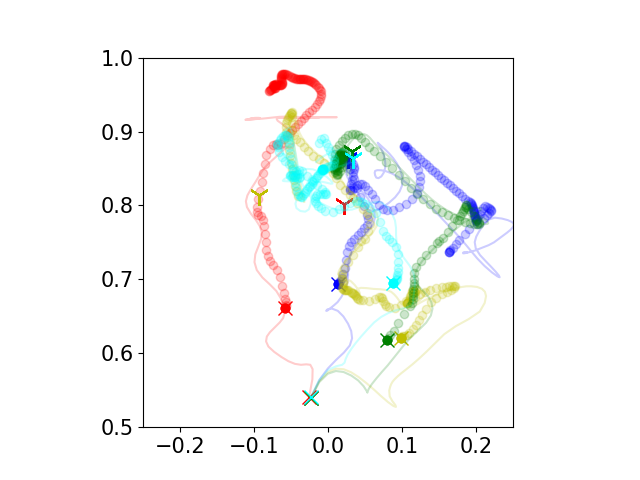}}
\end{subfigmatrix}
\caption[]{The center positions of the robot arm's fingers and the positions of objects in pick-place. The former and the latter are depicted as lines and circles. The initial positions are marked as ``x'', and the goals are marked as ``y''.
}
  \label{fig:trajml1}
\end{figure*}

To see what happened in meta-testing in pick-place, we analyze the trajectories of the center of the robot arm's fingers and the objects in the tasks.
The results are shown in \Figref{fig:trajml1}.
The experimental settings were the same as that of \Figref{fig:inference}.
\Figref{fig:trajml1} show that PEARL did not move the object in the first episode in some runs, while ELUE moved the object in all runs.
Note that transferring learned policies is difficult when there are only a small number of meta-train tasks.  
That is because in such a situation, the probability that a meta-test task is not similar to any of the meta-train tasks is high.
In our ML1 experiment, there were only 20 meta-train tasks,
this may be one of the reasons for the difficulty of transferring policies to the meta-test task.
\subsection{Learning Policies Without Embedding}
\begin{figure*}[tbh]
  \begin{subfigmatrix}{2}
 \subfigure[Third episode reward in pick-place]{
   \includegraphics[width=0.48\hsize]{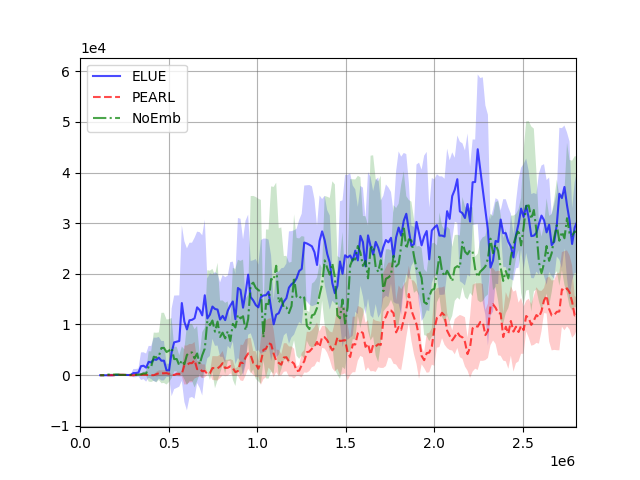}}
 \subfigure[Learning curves]{
   \includegraphics[width=0.48\hsize]{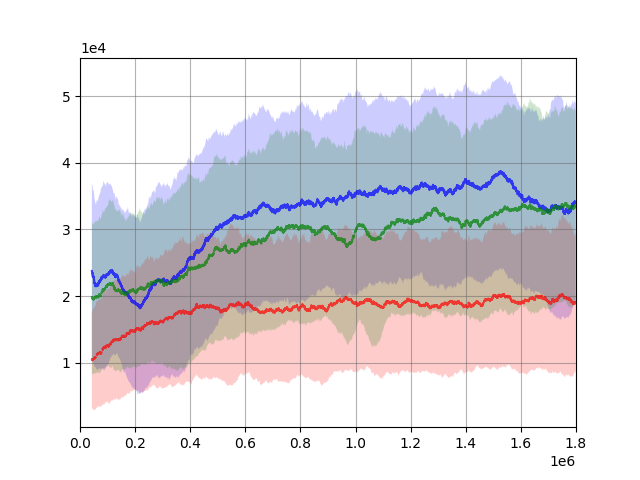}}
 \subfigure[Third episode reward in ML10]{
   \includegraphics[width=0.48\hsize]{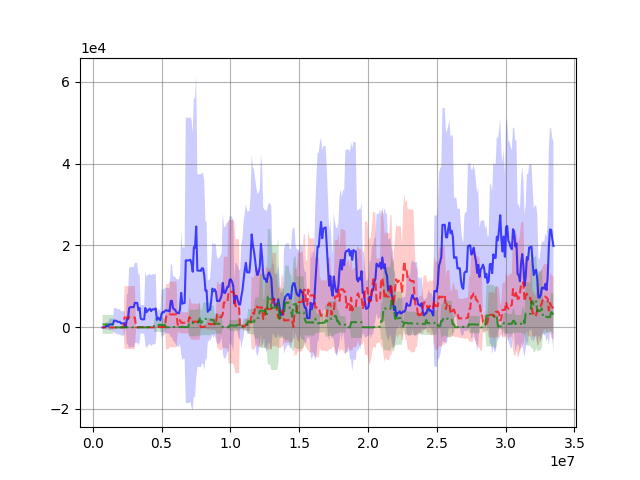}}
 \subfigure[Learning curves]{
   \includegraphics[width=0.48\hsize]{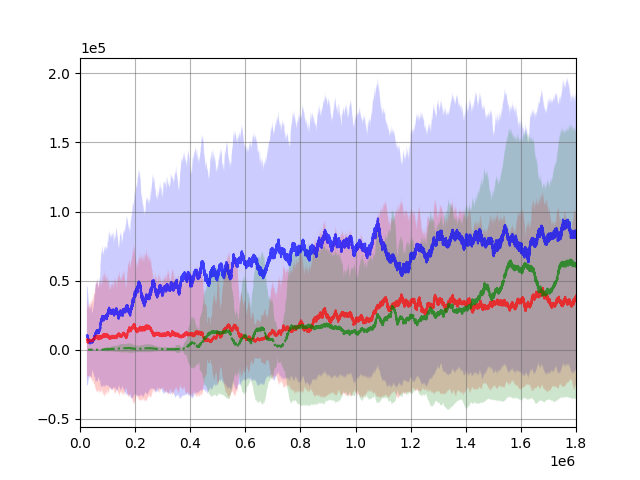}}
   \end{subfigmatrix}
   \caption[]{Comparison to learning without task embedding in ML1 pick-place and ML10.  The vertical axis is the average episode rewards in meta-testing and the horizontal axis is the number of time steps (a), (c) in meta-training and (b), (d) in meta-testing. }
  \label{fig:noemb}
\end{figure*}

For an ablation study, we examine the performance of ELUE without the embedding learning (i.e. SAC with the information bottleneck objectives), which is referred to as ``NoEmb''.
NoEmb does not have the ability to identify what the current task is.
This can be a drawback, e.g. each task requires contradicting actions at the same state.
After meta-training in pick-place and ML10, we executed meta-testing for each benchmark.
The experimental settings were the same as those of the former experiments in ML1 and ML10.
The results are shown in \Figref{fig:noemb}.
For comparison, the other results are also shown in \Figref{fig:noemb}.
To see the performance at the early stage, we show the third episode reward instead of the first one because the performance of PEARL was better as shown in \Figref{fig:validation0_ml1} and~\ref{fig:validation2_ml1}.
The results show that the performance of ELUE improved faster than that of NoEmb.
In ML10, the third episode reward of ELUE was better than that of NoEmb, while there was not a clear difference in the performance in ML1.
The results imply that a method without task information like NoEmb is one of the competitive baselines in some benchmarks, although it was not analyzed in some existing work~\cite{rakelly2019efficient, zintgraf2020varibad}.

\subsection{Conventional Benchmarks}
\begin{figure*}[tbh]
  \begin{subfigmatrix}{2}
 \subfigure[First episode reward in ant-fwd-back]{
   \includegraphics[width=0.48\hsize]{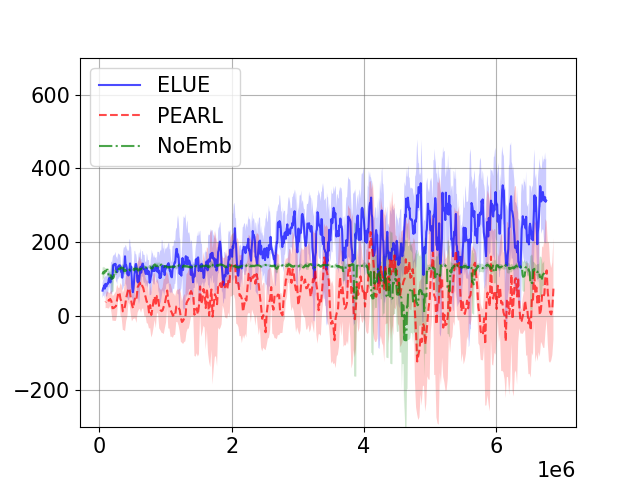}}
 \subfigure[Third episode reward in ant-fwd-back]{
   \includegraphics[width=0.48\hsize]{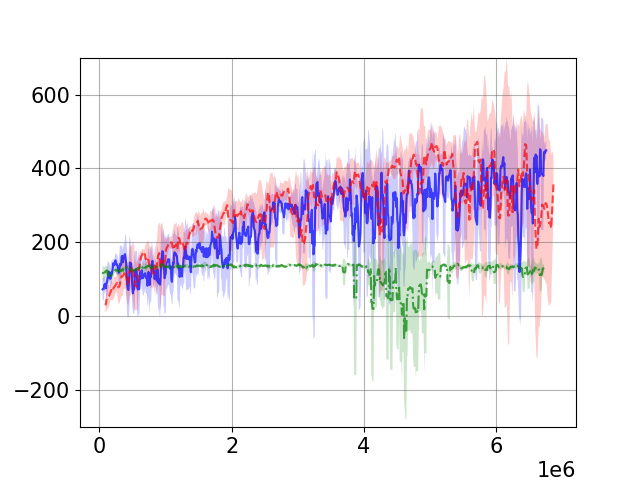}}
 \subfigure[First episode reward in humanoid-dir]{
   \includegraphics[width=0.48\hsize]{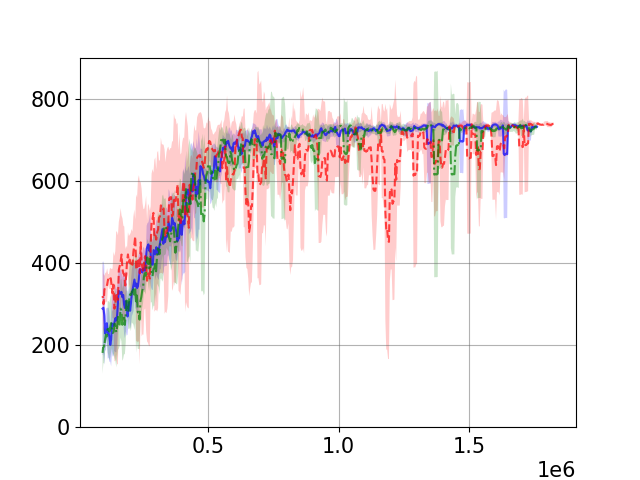}}
 \subfigure[Third episode reward in humanoid-dir]{
   \includegraphics[width=0.48\hsize]{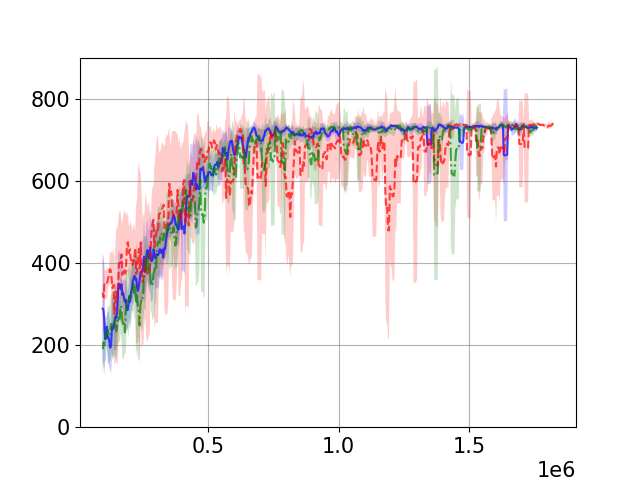}}
   \end{subfigmatrix}
   \caption[]{Comparison of sample efficiency of each method in the conventional benchmarks.  The vertical axis is the average episode rewards in meta-testing and the horizontal axis is the number of time steps in meta-training.}
  \label{fig:conventional}
\end{figure*}

We examine the sample efficiency of our method in the conventional benchmarks, ant-fwd-back and humanoid-dir, which are extensions of MuJoCo robot control tasks in OpenAI gym.
In the benchmarks, tasks differ in terms of goal directions of the ant robot or the humanoid robot.
In the tasks, their episode rewards are determined by the agent speed in the goal direction, alive bonus, and the other cost e.g. control cost.
The alive bonus is bonus of the agent is ``alive'' e.g. not falling and the control cost is cost of the agent to execute actions. 
In ant-fwd-back, there are only two tasks, tasks with forward and backward directions.
In these benchmarks, unlike Meta-World, the episode ends when the agent falls even before 150 steps.
The results are shown in~\Figref{fig:conventional}.

As for the ant benchmark, the performance of NoEmb did not improve more than about 150, while that of PEARL and ELUE were gradually improved.
Although the learning speed of ELUE was slightly worse than those of PEARL in the third episode reward,
the final performance of ELUE was slightly better.
In the first episode, ELUE achieved better results like that in ML1 benchmarks.

As for the humanoid benchmark, the first and third episode rewards were almost the same among the methods and their episode rewards were about 750.
It seems that the inference was not helpful for improvement of their performances. 
In this benchmark, the alive bonus was five.
Thus, if the agent is alive in a whole episode, its episode reward is $5 \times 150$ $+$ goal direction bonus $-$ control cost.
These results imply that most of the reward came from the alive bonus and that the goal direction bonus may be too small to learn different behaviors from task to task.

\comm{
  仮説として，「十分なゴールボーナスがあるが，control costで打ち消されている」というものに対して反論出来るか？　control costを見ていない以上反論不能？　
  弱反論１．「きれいに打ち消されるのは不自然」　
  弱反論２．「打ち消される時点で十分ではない」　
  仮説として，「「７５０前後の報酬はほとんどalive bonus 由来」ではなく，goal direction bonusによるもの」というものに対して反論出来るか？　control costを見ていない以上反論不能？　
  弱反論１．「きれいに７５０前後にcost等で打ち消されるのは不自然」　
  弱反論２．「もしgoal direction bonusが高いなら，それなりにtask to taskなbehaviorを学習出来ているはずで，task情報が改善に貢献しないのはおかしい」　
  In the experiments, the maximum number of steps in an episode is 150 (the same as Meta-World), while 600 in the original experiments~\cite{rakelly2019efficient}.\memo{check the fact}
In both benchmarks, 4 times the episode rewards in our results was better than the results in \cite{rakelly2019efficient}.
These results imply that our choice of hyper parameters were appropriate at least in these conventional benchmarks.} 

\comm{
%
いずれも，ほとんど差がつかない

antは，ELUEの方がepisode1では良いものの，episode 3では若干悪い，

humanoidはPEARLもepisode1から高い
PEARLが，最初から良いのは，単に，
alive bonus is 5 and it is relatively high,
most part of reward came from alive bonus....
750 - control cost + goal direction bonus
}

\subsection{Theoretical Supplement}
For any random variables $X,Y,Z$, (and their realization $x,y,z$,) and variational distribution $q$,
we show $I(X;Y|Z) \leq \E\left[\log \frac{p(X|Y,Z)}{q(X|Z)}\right]$, as with inequation (14) in \citet{alemi2016deep}.
\begin{align}
  I(X;Y|Z)=\E\left[\log \frac{p(X|Y,Z)}{p(X|Z)}\right]
\end{align}
and KL divergence $D_{KL}(p(\cdot|z)||q(\cdot|z)) \geq 0$.
Thus, 
\begin{align}
  &\E[- \log p(X|Z)]\leq \E[-\log q(X|Z)]\\
  &\iint p(z) p(x|z)\log \frac{1}{p(x|z)} dx dz \nonumber \\
  &\leq \iint p(z) p(x|z) \log \frac{1}{q(x|z)}.
\end{align}
Therefore, $I(X;Y|Z)$ is 
\begin{align}
  &\E\left[\log \frac{p(X|Y,Z)}{p(X|Z)}\right]\\
  &=\iiint p(z)p(x, y|z)\log \frac{p(x|y,z)}{p(x|z)} dx dy dz\\
  &=\iiint p(z)p(x|z)p(y | x,z)\log \frac{p(x|y,z)}{p(x|z)} dx dy dz\\
  &\leq \iiint p(z)p(x|z)p(y | x,z)\log \frac{p(x|y,z)}{q(x|z)} dx dy dz\\
  &=\E\left[\log \frac{p(X|Y,Z)}{q(X|Z)}\right].
\end{align}

\end{document}